\documentclass[letterpaper]{article} 
\usepackage{aaai25}  
\usepackage{times}  
\usepackage{helvet}  
\usepackage{courier}  
\usepackage[hyphens]{url}  
\usepackage{graphicx} 
\usepackage{natbib}  
\usepackage{caption} 
\usepackage{listings}

\usepackage[nolist]{acronym}

\usepackage{adjustbox}
\usepackage{algorithm2e}
\usepackage{amsmath}
\usepackage{amssymb}
\usepackage{array, makecell}
\usepackage{enumitem}
\usepackage{pifont}
\usepackage{tabularx}
\usepackage{dialogue}

\newcommand{\cmark}{\ding{51}}%
\newcommand{\xmark}{\ding{55}}%

\title{Towards Zero-Shot, Controllable Dialog Planning with LLMs}

\author{
    Dirk Väth\textsuperscript{\rm 1},
    Ngoc Thang Vu\textsuperscript{\rm 1},
}
\affiliations{
    \textsuperscript{\rm 1}University of Stuttgart\\
    dirk.vaeth@ims.uni-stuttgart.de, thang.vu@ims.uni-stuttgart.de
}

\begin{document}

\begin{acronym}
    \acro{LLM}[LLM]{Large Language Model}
    \acro{RL}[RL]{Reinforcement Learning}
    \acro{QA}[QA]{Question Answering}
    \acro{CTS}[CTS]{Conversational Tree Search}
    \acro{NER}[NER]{Named Entity Recognition}
    \acro{NLU}[NLU]{Natural Language Understanding}
    \acro{DST}[DST]{Dialog State Tracker}
\end{acronym}

\maketitle

\begin{abstract}

Recently, \acp{LLM} have emerged as an alternative to training task-specific dialog agents, due to their broad reasoning capabilities and performance in zero-shot learning scenarios.
However, many \ac{LLM}-based dialog systems fall short in planning towards an overarching dialog goal and therefore cannot steer the conversation appropriately.
Furthermore, these models struggle with hallucination, making them unsuitable for information access in sensitive domains, such as legal or medical domains, where correctness of information given to users is critical.
The recently introduced task \ac{CTS} proposes the use of dialog graphs to avoid hallucination in sensitive domains, however, state-of-the-art agents are \ac{RL} based and require long training times, despite excelling at dialog strategy.
This paper introduces a novel zero-shot method for controllable \ac{CTS} agents, where \acp{LLM} guide the dialog planning through domain graphs by searching and pruning relevant graph nodes based on user interaction preferences.
We show that these agents significantly outperform state-of-the-art \ac{CTS} agents ($p<0.0001$; Barnard Exact test) in simulation.
This generalizes to all available \ac{CTS} domains.
Finally, we perform user evaluation to test the agent's performance in the wild, showing that our policy significantly ($p<0.05$; Barnard Exact) improves task-success compared to the state-of-the-art \ac{RL}-based \ac{CTS} agent.

\end{abstract}

\begin{links}
    \link{Code}{https://github.com/DigitalPhonetics/conversational-tree-search/tree/llm-policy}
\end{links}

\section{Introduction}
Recently, \acp{LLM} have emerged as an alternative to training task-specific dialog agents.
They excel in broad reasoning capabilities \cite{huangReasoningLargeLanguage2023}, and demonstrate high performance in zero-shot learning scenarios \cite{wei2021finetuned, brown2020language}.
However, \acp{LLM}-based dialog agents often fall short when planning towards a dialog goal is required.
As the underlying \acp{LLM} are optimized towards following user instructions, such agents struggle to take initiative in proactively steering dialogs towards a fixed goal across multiple turns \citep{deng2024plugandplay}.
Additionally, these models struggle with hallucination \cite{xuHallucinationInevitableInnate2024,liDawnDarkEmpirical2024}, making them unsuitable for information access in sensitive domains where the information given to a user must be correct.

Because of these challenges, common language-based interfaces for accessing information in sensitive domains rely on FAQ-style retrieval or hand-crafted dialog policies.
However, while FAQ-style retrieval agents can deliver immediate answers, they are limited to general information not tailored to a users specific situation, which leads to either very general or overly long and complex responses.
Furthermore, retrieval accuracy can suffer in large domains.
Dialog systems with hand-crafted policies, on the other hand, can provide short and personalized responses, but are often perceived as frustratingly long and rigid.

\begin{figure}[tb]
    \centering
    \includegraphics[width=0.85\linewidth]{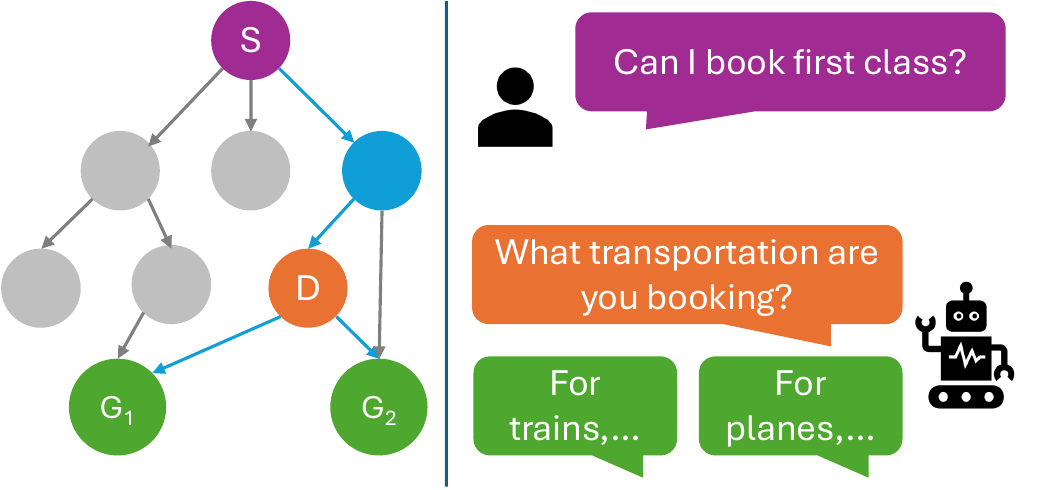}
    \caption{To plan a dialog, the proposed agent tracks possible user goals (green: $G_1, G_2$) based on the initial user input (purple: $S$), then plans paths (blue: $S \rightarrow ... \rightarrow D \rightarrow G_{1/2}$) to reach them, identifying decision points to proactively prompt the user for more information (orange: $D$).}
    \label{fig:teaser}
\end{figure}

The recently introduced task, \ac{CTS} \cite{vathConversationalTreeSearch2023}, proposes a graph-based framework for controllable task-oriented dialog in sensitive domains that interpolates between these two information seeking styles, implicitly adapting its behavior to match user interaction expectations.
This flexibility allows \ac{CTS} to combine the advantages of both settings, while avoiding their largest disadvantages.
To prevent the policy from switching contexts or hallucinating false information, \ac{CTS} requires a dialog agent to walk an expert-created dialog graph, deciding at each node whether it is relevant to the user and should be output, or which neighbor to skip to next.
This graph walking approach requires the ability to plan efficient paths to possible goal nodes and to identify the decision points where additional user input is necessary.
As \ac{RL}-based algorithms are trained to optimize goal planning,
\citet{vathConversationalTreeSearch2023, vathZeroDataControllableAdaptive2024} propose \ac{RL}-based agents for the \ac{CTS} task.

Despite these advantages, a big drawback of the aforementioned \ac{RL}-based state-of-the-art \ac{CTS} agents is their long training time (up to 7 days on GPU for a graph with 123 nodes).
This is particularly problematic as a new agent must be trained every time the graph is updated, even if only a single node changes, making it difficult to develop new dialog domains in an iterative way.
Not only does this have a big impact on the environment and costs, but in sensitive domains such as legal domains, it could be unacceptable to have such long waiting times in cases where new laws or policies come into effect.

The goal of this paper is therefore to explore how to apply the reasoning and zero-shot capabilities of \acp{LLM} to dialog planning in a way that allows steering the conversation towards an overarching goal, while at the same time avoiding the hallucination problem and maintaining an expert-controllable dialog flow. 
To this end, we investigate the following research questions:
\begin{itemize}
    \item \textbf{RQ1:} 
    How can we optimize the runtime efficiency (to allow for real-time dialogs) and resource requirements of a zero-shot \ac{LLM}-based \ac{CTS} agent when planning dialog trajectories through a \ac{CTS} domain graph?
    \item \textbf{RQ2:}
    How can we optimize the dialog success of such a zero-shot \ac{LLM}-based \ac{CTS} agent?
    \item \textbf{RQ3:} How does a zero-shot \ac{LLM}-based approach to dialog planning compare to a trained \ac{CTS} \ac{RL}-agent in simulation?
    \item \textbf{RQ4:} How does a zero-shot \ac{LLM}-based approach to dialog planning compare to a trained \ac{CTS} \ac{RL}-agent when testing with real users?
\end{itemize}

To address these questions, we investigate how \acp{LLM} can be leveraged to help plan conversations by first detecting the user preference for interaction style, searching the domain graph for relevant nodes, pruning the search results, and subsequently guiding the user across selected paths through the domain graph (see figure \ref{fig:teaser}), while at the same time eliminating the need for training and adhering to the \ac{CTS} controllability aspects.
We validate our method against state-of-the-art \ac{RL}-based agents in simulation on multiple domains.
Finally, we evaluate real-world performance in a user study. 

Our main contributions are:
1) We propose a novel, \ac{LLM}-based method for a zero-shot, controllable task-oriented \ac{CTS} agent: \ac{CTS}-LLM.
2) We demonstrate strategies to improve the reasoning quality and inference speed of this \ac{LLM}-based agent, making real-time dialog feasible.
3) We show that \ac{CTS}-LLM is not only able to successfully plan and execute dialogs, but even improves dialog success in simulation compared to trained state-of-the-art \ac{RL}-based \ac{CTS} agents.
These results generalize to all three available domains, when using large commercial as well as smaller scale local \acp{LLM}.
4) Finally, we demonstrate that improvements to dialog success translate to the real world through human evaluation on a single domain. 
All code and data are publicly available.

\section{Related Work}

In this section, we give a brief overview of the most closely related methods. 

\subsection{Dialog Planning}

In task-oriented dialog systems, a policy, sometimes also called dialog manager, is responsible for selecting the dialog system's next actions.
There are many different ways to implement dialog policies, such as using a fixed set of rules \citep{gus}, statistical models \citep{kim2008frame}, or even prompting or fine-tuning \acp{LLM} \citep{mi2022cins}.

One of the most frequently applied approaches is to train task-oriented dialog policies with \ac{RL} because such agents naturally learn efficient planning strategies to reach user goals \citep{scheffler2002automatic, peng-etal-2018-deep, kwan2023survey}.
These \ac{RL} agents are usually trained in an interactive setting, by performing dialogs against a simulated user \citep{schatzmann2006}.
Each dialog, the user simulator randomly selects a goal and then takes turns conversing with the \ac{RL} agent, until either reaching this goal, or encountering another constraint such as maximum dialog length.
While \ac{RL} agents perform well in task-oriented dialog, a new dialog agent must be trained for every new dialog domain.

\subsection{Conversational Tree Search}
\ac{CTS} \citep{vathConversationalTreeSearch2023} is a task-oriented, controllable dialog task which frames dialog planning as a graph search and traversal problem.

The graph, which describes the dialog domain, is structured by a human domain expert.
Each node contains text that the dialog agent can output to the user, e.g. a greeting, information, or follow-up questions.
As the possible output texts are defined by a human dialog designer, no hallucination is possible. 
Nodes are connected to neighbors via edges, where each edge is associated with a specific user intent, e.g., indicating interest in reimbursement or in booking.
Every node is associated with a node type, such as question and variable nodes (for acquiring user input), logic nodes (choosing neighbor nodes based on variable values instead of user intents), and information nodes (information output without user input).

A \ac{CTS} agent walks this graph node by node, starting from a unique greeting node.
The goal is to identify and reach the node in the graph that best answers the user's information need.
At each node, the agent makes the decision to either output the current node's text (e.g., asking a question or providing information), or to skip a neighboring node.

\begin{figure}
    \centering
    \includegraphics[width=0.45\textwidth]{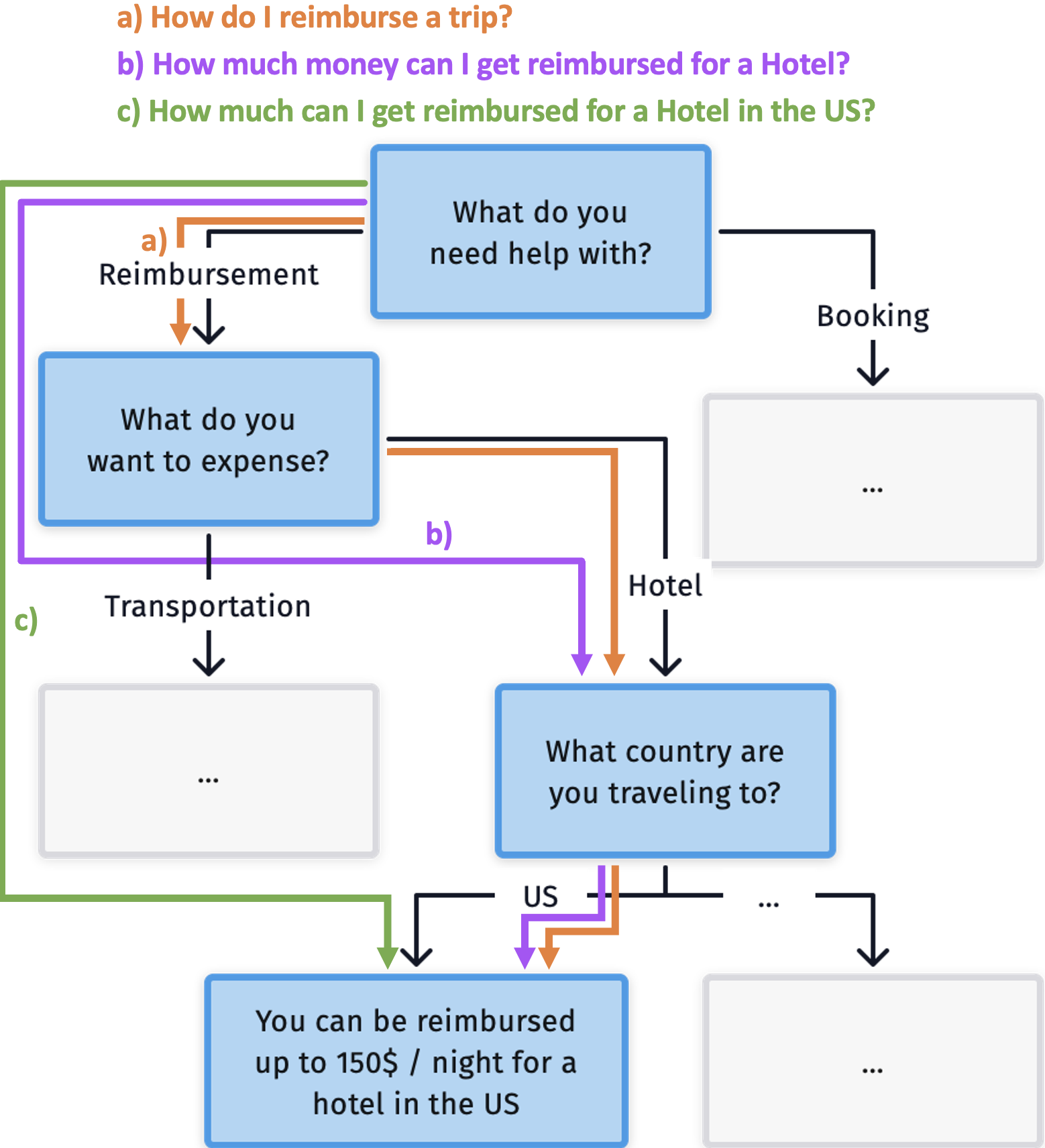}
    \caption{Example dialog graph and three different user inputs, leading to step-by-step conversation (a), clarification steps (b), or directly skipping to the answer (c) \citep{vathConversationalTreeSearch2023}}
    \label{fig:task_example}
\end{figure}

The task supports different user interaction preferences, accommodating both users with a vague or general information need, as well as users with more concrete questions.
Based on the predicted interaction style, an agent can either traverse the graph by 1) follow a rigid scheme where the system outputs the text of each node visited, e.g., if the user information need is very vague, 2) skip directly to an answer, e.g., if the user question is very concrete, or 3) employ a hybrid approach, 
asking only questions that are necessary to clarify the user's information need on the path to the nodes that are candidates for potentially providing the requested information.
Figure \ref{fig:task_example} shows an example of three different user inputs, and the three different resulting interactions.
As the agent can only output predefined text or filled templates, the output information is controllable by the dialog designer.
Regardless of interactions style, the agent must walk the dialog tree node by node, meaning it cannot suddenly switch contexts.

Current state-of-the art \ac{CTS} agents are \ac{RL}-based \citep{vathConversationalTreeSearch2023, vathZeroDataControllableAdaptive2024}, and while demonstrating good task performance, require long training times if any portion of the graph is changed.
To this end, our goal is to replace the \ac{RL}-based agent with one that can zero-shot adapt to any graph changes or even new graphs.

\subsection{Zero-Shot Learning}
Recent research into \acp{LLM} has demonstrated their strong abilities as zero-shot learners \citep{NEURIPS2023_3eb7ca52}.
This performance can be improved by fine-tuning models on tasks which are described by textual instructions \citep{wei2021finetuned}.
\citet{brown2020language} further show that similar improvements can be obtained by providing in-context examples.
Rather than providing only instructions for the task the model should accomplish, the authors additionally included examples of how to do the task in their prompt.
Furthermore, \citet{schick-schutze-2021-just} found that in-context examples are not just useful for very large \acp{LLM}, but can also be used to improve the performance of smaller, less resource intensive language models.

\subsection{Information Retrieval}
As this paper focuses on information seeking scenarios, we also discuss related work in the realm of information retrieval.
Although there has recently been a shift towards retrieval augmented generation \citep{NEURIPS2020_6b493230, Chen_Lin_Han_Sun_2024}, these methods can still suffer from the same hallucination problems inherent to \acp{LLM} \citep{10.1093/jamia/ocae014}, making them unsuitable for sensitive domains.
Therefore, we primarily discuss document based retrieval,
which has long been an area of research \citep{Mitra2000, Ghorab2013, 10.1145/3648471}.
Early methods relied on co-occurrences of words or lemmas between the search query and the available documents \citep{salton1983introduction}. 
Modern approaches tend to make use language embeddings that are better able to represent the meaning of documents \citep{10.1145/3648471}, sometimes adding a filtering or re-ranking step to the retrieved documents \citep{zerveas-etal-2022-coder}.
Our proposed method also makes use of such embedding models for the initial document retrieval, and an \ac{LLM} for a filtering step.

\subsection{Natural Language Understanding} 
Natural language understanding has been a long-standing goal in the NLP community \cite{SCHANK1972552, 10.1145/3547138, 10188908}.
Although many approaches have been applied over the years, from handcrafted rules \cite{relax_nlu}, to probabilistic models \citep{391397, 1318505}, to deep learning \citep{6707709}, more recent approaches have begun leveraging the power of \acp{LLM} \citep{10.1145/3605943}.
In this paradigm, rather than needing to learn a mapping from user utterances to a fixed set of slots and intents, the problem can be reformulated as a question answering task \citep{9413810}.
Here, the goal is to leverage the knowledge and language understanding capabilities of language models to answer the question of ``which intent/slot is the user trying to express?'', and by doing so, to reduce or eliminate the need for training data.

Our method similarly uses \acp{LLM} to match user inputs against the set of possible intents (edges) at a given dialog node, following the most similar intent (edge). 

\section{Method}

\begin{figure*}[htb]
    \centering
    \includegraphics[width=0.7\textwidth]{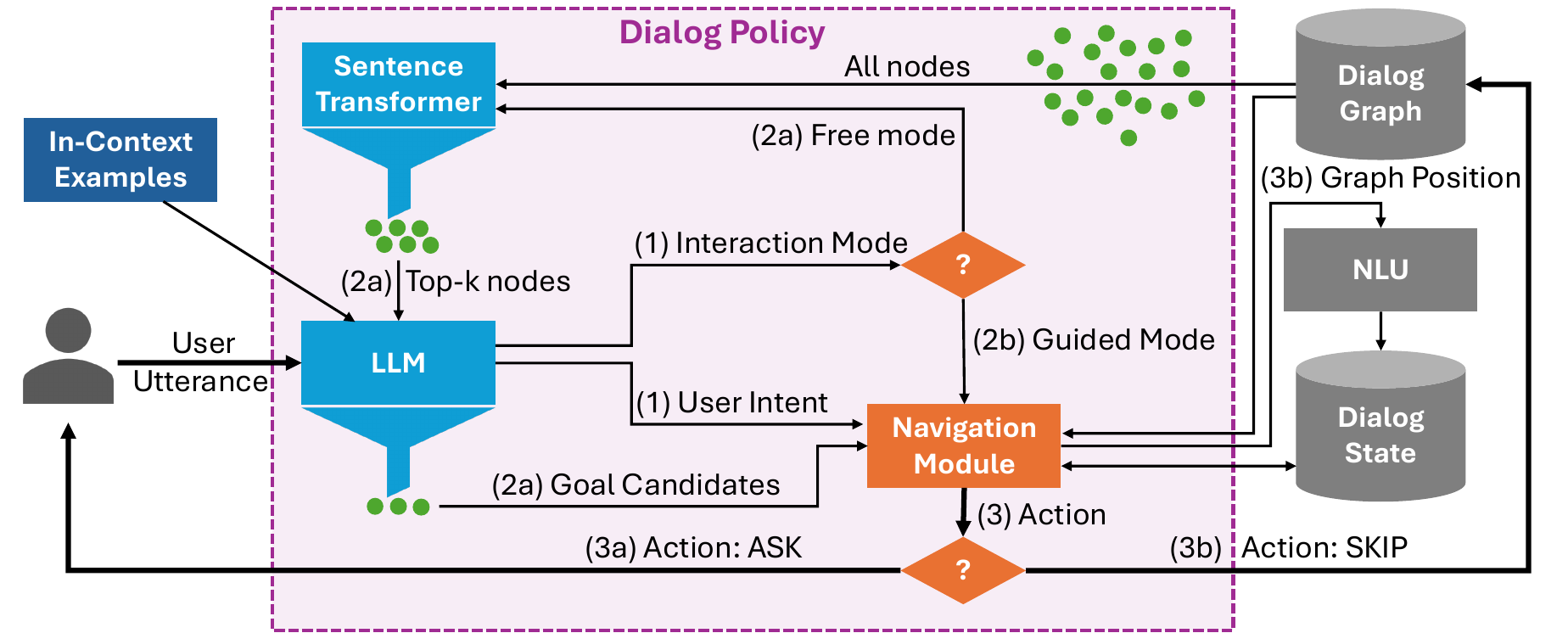}
    \caption{Interaction between user and system. The system predicts user intent and interaction mode from the user utterance (1). If the interaction mode is \emph{free}, the user utterance is used to retrieve answer candidates from the dialog graph (2a). In \emph{guided} mode (2b), the policy follows the edge associated with the predicted user intent (3b). The agent may also decide to ask for clarification or output information (3a).}
    \label{fig:method:system}
\end{figure*}

Our proposed \ac{LLM} dialog system is a module based system (see figure \ref{fig:method:system}) with the following components: 
1) a dialog policy, 
2) a \ac{NLU} unit, and
3) a dialog state tracker, storing variable values to fill templates.
The dialog policy consists of 
a) an \ac{LLM}-based interaction mode classifier, which decides whether the user inputs a statement or a question,
b) an \ac{LLM}-based intent classifier, to determine per node which intent, i.e., edge, a user was trying to select,
c) an \ac{LLM}-based filter module, searching node candidates in the dialog graph that might answer a user's question, 
and d) a navigation module responsible for walking the dialog graph and choosing to output or skip a node,

The policy adheres to the \ac{CTS} task by traversing the dialog graph node by node, only outputting predefined node texts, thus remaining controllable.
Even though we use \acp{LLM} to make decisions in how to traverse the graph, their outputs are never shown to the user.

\subsection{Dialog Policy}

In contrast to \ac{RL}-based agents, the dialog planning in our method is performed by the combination of an \ac{LLM} and graph algorithms.
Concretely, the dialog policy uses the same \ac{LLM} for the components a), b) and c) to coordinate planning (see algorithm  \ref{alg:policy} for a concise overview).

\subsubsection{Navigation Module}

The policy initiates a dialog by outputting the start node of the dialog graph to the user.
After receiving the first input from the user, the interaction mode classifier (step 1 in figure \ref{fig:method:system}) is called to decide whether the user expresses a vague information need, or poses a concrete question, then proceeding with either step:

\textbf{(2a)} If the user input was classified as a vague or general information need, the following dialog interaction is considered a ``guided" turn-by-turn mode with the dialog system in the lead, supporting users unfamiliar with the domain or with a vague information need.
That is, the policy will always output the content of the current node to the user and, if necessary, wait for user input to continue (step 3a in figure \ref{fig:method:system}).
The user's input will be matched against the possible intents available at that node (step 2a in figure \ref{fig:method:system}).
Then, the policy will move to the corresponding neighboring node (step 3b in figure \ref{fig:method:system}).
Dialog continues in this manner until the tree end is reached or the user ends the dialog.

\textbf{(2b)} If the user input was classified as a (concrete) question, the following dialog interaction is considered a ``free" mode, with the goal to answer the user's question as efficiently as possible, only prompting the user when disambiguation is required to clarify the exact information need.
From the dialog graph $G=(V,E)$, with nodes $V$ and edges $E$,
a list of possible goal node candidates $V_g \subseteq V$ is retrieved by the filter module (step 2b in figure \ref{fig:method:system}).
A node is considered a goal candidate if it might answer the user question.
For each goal node candidate $v_g \in V_g$, the policy calculates all possible paths $P_g := \{(v_1, \dots, v_n) | v_n=v_g \land \forall v_i, v_{i+1}: (v_i, v_{i+1}) \in E\}$ in the dialog graph that lead from the current node $v_1 \in V$ to that goal node.
Between the list of paths to each goal node, it selects the longest shared path prefix reaching all goal nodes.
Here, we calculate the shared prefixes by looking at all paths to the candidate goals: $P_{shared} := \{ (v_1, \dots, v_n) | \forall v_g \in V_g \exists p \in P_g \forall 1 \leq i \leq n: p_i = v_i \}$.
Once the longest prefix $argmax_{p \in P_{shared}} |p|$ has been identified, the policy walks through this prefix without outputting any of the nodes to the user until it reaches the final node (see step 3b in figure \ref{fig:method:system}).
The final node in the prefix is then output to the user (step 3a in figure \ref{fig:method:system}), because it is by definition either a decision point where user input is required, or a goal node itself.
If there are still open goals, the user intent is identified, the policy moves to the corresponding neighbor, and a new longest prefix is calculated using the remaining goal nodes.
We use the longest shared path prefix as users in our pilot study perceived this approach as more usable, e.g., compared to using shortest paths, as the decisions at the tail of the paths seem more related to the user question (closer to the goal) than at the beginning of the paths, where topics might still be quite general (farther removed from the goal).

As variable questions do not cause splits in the graph, they are skipped.
Whenever the policy reaches a node containing a template, but does not know the necessary values from the \ac{DST} yet, it looks back through the history of visited nodes until if finds the corresponding variable node. 
The policy then requests the variable value from the user, updates the beliefstate, and continues the dialog from the template node.

The full planning process is visualized in figure \ref{fig:teaser}, where the policy calculates paths (blue) to two possible goal candidates (green), asking for clarification only at the orange node (last node on longest shared path prefix) which is directly connected to the two possible goal candidates, rather than at preceding branching points (blue).

\RestyleAlgo{ruled} 
\SetKwComment{Comment}{/* }{ */}
\begin{algorithm}[h]
\caption{Dialog Policy}
\label{alg:policy}
\KwData{$graph, input, node$}

$done \gets false$ \;
$mode \gets \text{mode\_classifier}(input)$\;
\If{$mode = \text{QUESTION}$}{
    $goals \gets \text{goal\_filter}(graph, input)$\;
    $path \gets \text{longest\_prefix}(goals)$\;
}
\While{$done \neq  true$}{
    \If{$mode = \text{QUESTION}$} {
        $node \gets \text{last\_node}(path)$\;
    }
    $\text{output}(node)$\;
    \uIf{$node.type \in \{\text{QUESTION, VARIABLE}\}$} {
        $input \gets \text{input()}$\;
        $next\_node \gets \text{neighbors}(node)[intent(input)]$\;
        $path \gets \text{longest\_prefix}(goals)$\;
    }
    \Else {
        $next\_node \gets \text{neighbor}(node)$\;
    }
    \If{$node \in goals$} {
        $goals \gets goals \setminus \{ node \}$\;
    }
    $node \gets next\_node$\;
    $done \gets goals = \emptyset \lor \text{neighbors}(node) = \emptyset$\;
}
\end{algorithm}

\subsubsection{Interaction Mode Classification}
\label{sec:intent_classifier}

Interaction mode classification is the first step in planning dialogs with our proposed method.
We classify the user utterance into the classes \emph{guided} or \emph{free} by prompting an LLM to determine if a given user input is a question/command/request or not (see appendix \ref{sec:llm_prompts} for the exact prompt).
We then parse the generated outputs \emph{`yes', `command', `request'} as \emph{free} and all other outputs as \emph{guided}. 
Using an \ac{LLM} for this task avoids training a specialized classifier and detects the interaction mode better than, e.g., off-the-shelf question classifiers, as many user inputs that are asking for information might not be phrased as questions.

\subsubsection{Intent Classification}

Each decision (question) node has a list of possible user intents.
To decide what intent the user is trying to express, we prompt an LLM with the user utterance and a JSON list containing the intent candidates and their indices. We ask the model to output the index of the response that best matches the user's input (see appendix \ref{sec:llm_prompts}).
We then extract this index and move to the corresponding neighbor.

\subsubsection{Goal Node Filter}

The goal node filtering step is responsible for the path selection aspect of dialog planning in \emph{free} policy mode.
We perform a similarity ranking between the user utterance and the list of all nodes in the graph using a fast semantic search model, retrieving the 15 most similar nodes as a pre-filtering step.
These nodes are then passed into an \ac{LLM} to reason about which candidates could actually be answers to the user's question. 
Performing the retrieval step with a much smaller model instead of giving the full list of nodes to the \ac{LLM} keeps the number of input nodes to the \ac{LLM} constant, regardless of graph size.
This reduces the number of input tokens and the search space to be reasoned over, thus simplifying the problem and reducing computation time, while also increasing the accuracy of the results.
However, fast retrieval methods usually obtain reasonable recall scores for $k >> 1$, and thus the top-1 result cannot be trusted (also demonstrated in figure \ref{fig:prefilter}).

We therefore apply \ac{LLM}-based reasoning to the 15 answer node candidates conditioned on user utterance to filter out the most relevant candidates regardless of their retrieval rank.
We then ask the model to output all nodes which answer the question, along with a justification for why each node does so. 
To improve the reasoning process, we add out-of-domain, in-context examples that represent positive and negative question/answer pairs.
For example, we add:
`\emph{What is the weather usually in Singapore at 9 a.m.?}', 
and as a positive example fact:
`\emph{In Singapore, between 8 a.m. and 11 a.m., the weather is around 35 degrees Celsius.}'
with the justification:
`\emph{The fact is relevant as it answers the user request, and the requested time of 9 a.m. lies between the fact's timespan of 8 a.m. to 11 a.m.}'.
The full prompt can be seen in appendix \ref{sec:llm_prompts}.

\section{Datasets}

To investigate how our method compares to state-of-the-art \ac{RL}-based \ac{CTS} agents, we evaluate on the three available English datasets: REIMBURSE-en, DIAGNOSE and ONBOARD.
REIMBURSE-en contains information about travel reimbursement procedures, DIAGNOSE is a medical help domain, and ONBOARD provides support for foreigners moving to a new city.

Each of these datasets contains a dialog graph, user questions and user responses.
User questions are associated with the dialog node that provides an answer.
User responses are paraphrases for the possible user intents at each node.
Statistics for each dataset can be found in table \ref{tab:datasets}.

\begin{table*}[htb]
    \center
        \begin{tabular}{|l|l|c|c|c|c|c|c|c|}
            \hline
            \textbf{Dataset} & \textbf{Split}  & \textbf{\#Nodes} & \textbf{\makecell{Tree\\Depth}} & \textbf{\makecell{Max. Node\\Degree}} & \textbf{\makecell{\#User\\Questions}} & \textbf{\makecell{Avg. User\\Questions}} &\textbf{ \makecell{\#Answer\\Paraphrases}} & \textbf{\makecell{Avg. Answer\\Paraphrases}}  \\ 
            \hline
            REIMBURSE-en & Train & 123 & 32 & 14 & 279 & 3.5 & 246 & 3.4 \\
            REIMBURSE-en & Test & 123 & 32 & 14 & 173 & 2.2 & 162 & 2.2 \\
            DIAGNOSE  & Test & 98 & 10 & 6 & 150 & 2.0 & 298 & 3.0 \\
            ONBOARD   & Test & 88  & 15 & 9 & 117 & 2.0 & 152 & 2.7 \\ 
            \hline
        \end{tabular}
    \caption{Overview of the \emph{REIMBURSE-en}, \emph{ONBOARD} and \emph{DIAGNOSE} datasets \cite{vathZeroDataControllableAdaptive2024}.}
    \label{tab:datasets}
\end{table*}

\section{User Study}
To assess the performance of our proposed CTS-LLM agent, we performed user evaluation.
To be able to compare our results to \citet{vathZeroDataControllableAdaptive2024}, we performed a parallel study adhering to the same design, methodology, and REIMBURSE-en domain.
Although we only evaluated the \ac{CTS}-\ac{LLM} agent, we can thus treat the combination of both studies as a between-subject design.

\subsection{Study Design}
All participants were recruited via the crowdsourcing platform Prolific and paid at a rate of 12.60€/hr, consistent with minimum wage in the country of our research institution.
They were asked to provide basic demographic information about their age, gender, and experience with chatbots and business travel.
Following \citet{vathZeroDataControllableAdaptive2024}, subjects were then asked to conduct three conversations with the chatbot in the REIMBURSE-en domain.
For each dialog they were randomly assigned an information goal they should try to reach.
These goals represented 1) an open question, e.g., the general process of how to book a hotel, or what to do in an emergency, 
2) a simple, concrete question, e.g., if they could get reimbursed for seat reservations on a train,
or 3) a complex, concrete question, e.g., how much money they could be reimbursed per day if they planned to stay at their brother's apartment on their upcoming trip to France.

After each dialog, users were asked to rate the subjective quality and length, using two single-item Likert questions.
Subjective length was rated on a 5-point scale from 1 (much too short) to 5 (much too long).
Success was rated on a 4-point scale from 1 (`my question was not answered at all') to 4 (`my question was entirely answered').
After the final dialog, participants were also asked to provide free-form feedback on their perception of the agent as well Likert feedback on the usability of the agent using the UMUX scale \citep{finstad2010usability}.
They were also asked about their trust in the agent and perceived reliability using the Trust and Reliability subscales of \citet{korber2018theoretical}'s Trust in Automation scale.

\subsection{Pilot Study}
Before conducting the main user study, we performed a pilot study with 12 users to check the technical setup.
Here, we found that users were quickly frustrated when they had to answer multiple questions at the beginning of the dialog tree, as they took this to mean that the agent had not understood their question. 
This motivated our design decision to delay clarifications by calculating the longest shared path prefix instead of shortest paths.

\subsection{Main Study}
For the main study, we recruited 24 participants.
Of these, 13 identified as female and 11 as male, with ages ranging from under 20 to 59 years old.
On average, participates had decent familiarity with chatbots (3.5 on a 5-point Likert scale) and some familiarity with business trips (2.5 on a 5-point Likert scale).
Each user participated in 3 dialogs, resulting in a total of 72 collected dialogs.
After removing 4 dialogs due to technical errors or bad-faith participation, we were left with 68 total dialogs.
Combining these with the 61 dialogs collected for \ac{CTS}-\ac{RL} by the parallel study from \citet{vathZeroDataControllableAdaptive2024} resulted in a total of 129 dialogs for the between-subject setup.

\section{Results \& Discussion}

All results for the \ac{LLM}-CTS agents were obtained using GPT-4o-mini (version 2024-07-18)\footnote{\url{https://platform.openai.com/docs/models/gpt-4o-mini}} or the instruction tuned Gemma-2 9B \cite{gemma_2024}.
We choose the multi-qa-mpnet-base-dot-v1 Sentence Transformer \cite{reimers-2019-sentence-bert} for pre-filtering.
Results for the \ac{RL}-based CTS agents are taken from \cite{vathZeroDataControllableAdaptive2024}.

\subsection{Objective Evaluation Metrics}
To evaluate the objective performance, we follow \citet{vathZeroDataControllableAdaptive2024} and measure binary dialog success by evaluating whether the user's information need was fulfilled for each dialog.
This means that the dialog agent has to reach the node containing the answer to the user's information goal as well as output the node text to the user.
The reported values for dialog success measure the percentage of successful dialogs.

Dialog length refers to the number of dialog turns (agent turns + user turns) which are perceived by the user.  
Thus, dialog length only counts the number of user inputs and the number of system \emph{ASK} actions.
\emph{SKIP} actions are not counted, as they are not visible to the user.

Finally, interaction mode classification is a binary classification measure (\emph{free} or \emph{guided}), measured once per dialog based on the prediction from the initial user utterance.

\begin{table*}[htb]  
\begin{tabular}{|l|c|c|c|c|c|c|c|c|}
\hline
\textbf{Pre-filter} & \textbf{Justifications} & \textbf{\makecell{In-\\Context\\ Examples}} & \textbf{Recall} & \textbf{\makecell{Avg.\\ Candidates\\per Question}} & \textbf{\makecell{Avg. time\\per Question \\(sec)}} & \textbf{\makecell{Avg.\\ $\#$tokens\\(input)}} & \textbf{\makecell{Avg.\\$\#$tokens\\(output)}} & \textbf{\makecell{Avg. GPU \\memory\\(GB)}} \\
\hline 

\cmark & \cmark & \cmark & \textbf{0.84} & 2.37 & 17.84 & 1581 & 95  & 26.59 \\
\cmark & \cmark & \xmark  & 0.35 & 0.91 & 25.85 & 935  & 147 & 26.49 \\
\cmark & \xmark  & \cmark & 0.80 & 2.26 & \textbf{6.66}  & 1346 & 31  & 26.56 \\
\xmark & \cmark & \cmark & 0.78 & 2.66 & 33.61 & 4601 & 158 & 27.04 \\ 
\hline
\end{tabular}
\centering

\caption{Recall and resource usage for the REIMBURSE-en train split with Gemma-2 on a NVIDIA A600 GPU for different goal candidate filtering processes.}
\label{tab:resources}
\end{table*}

\subsection{RQ1: Optimizing Resource Requirements \& Runtime Efficiency}

To save on hardware requirements, we try to solve as many of the policy sub-tasks as possible with the same \ac{LLM}.
We used Gemma-2 on the train split of the REIMBURSE-en dataset as a development setup.

After conducting preliminary experiments, we find that we can obtain good performance from the same \ac{LLM} for interaction mode detection, user intent classification, and goal node candidate filtering.
However, as can be seen from table \ref{tab:resources}, filtering goal node candidates directly from the full dialog graph proved too slow for real-time usage.
In our experiments, we found that this approach to filtering takes $\sim$33.61 seconds, and is resource intensive because of the number of resulting input and output tokens (4601 and 158, respectively). 
This will only get worse as the graph size increases.

\begin{figure}
    \centering
    \includegraphics[width=\linewidth]{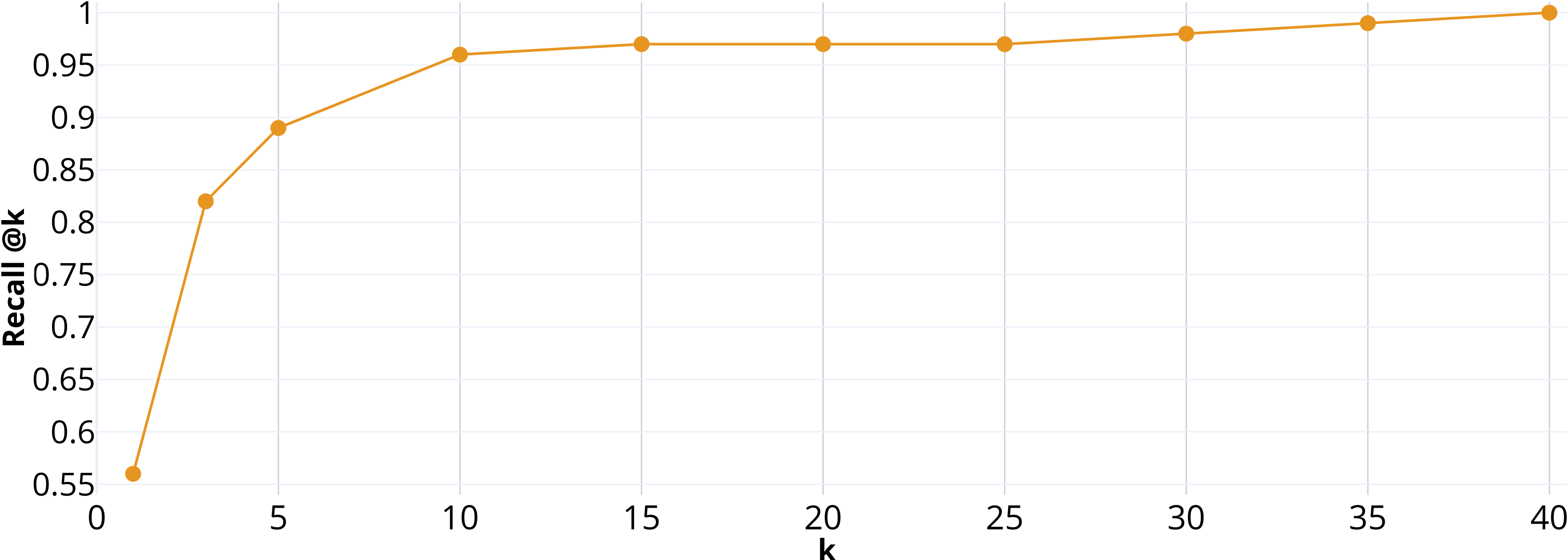}
    \caption{Recall of pre-filtering the possible goal nodes for different $k$ on the REIMBURSE-en train split.}
    \label{fig:prefilter}
\end{figure}

To address this, we add a small Sentence Transformer specialized in relevant passage retrieval \cite{reimers-2019-sentence-bert} to pre-filter nodes based on semantic similarity to the user utterance.
After performing a retrieval experiment on the REIMBURSE-en train split (see figure \ref{fig:prefilter}), we choose $k=15$ as a good trade-off between candidate space reduction and accuracy.

Keeping all 15 goal candidate nodes, however, would result in the need for many clarifying questions and thus long dialogs.
To mitigate this, we choose to apply the \ac{LLM} as a post-filter to the 15 retrieved candidates.
Together, the retrieval and post-filtering process only take $\sim$17.84 seconds, almost halving the execution time, making it feasible for real-time use.
Since the retrieval step is very efficient, it scales better with graph size while the slower post-filter is now decoupled from the graph size.
This allows for scaling to larger domains without a noticeable penalty on filter time.
As we see in table \ref{tab:resources}, even though we add the additional retrieval model, the reduction in input (factor $\sim$2.9) and output (factor $\sim$1.7) tokens results in a lower total memory footprint: 26.59 GB, compared to 27.04 GB for the \ac{LLM} without the retrieval step.
Finally, the reduction in tokens and computation time also lowers the environmental impact, and especially for \acp{LLM} consumed via API, costs.

\subsection{RQ2: Optimizing Dialog Success}

Performance of \emph{free} mode is fully dependent on the output quality of the filtering step, thus improving filter performance directly results in better dialog success.
Again, we use Gemma-2 on the train split of the REIMBURSE-en dataset as a development setup.

As first tests with the two-stage filtering process did not yield good retrieval accuracy, we analyzed filtering failures.
We observed, e.g., that some \acp{LLM} have trouble with numerical reasoning.
For example, given the user question: `\emph{How much money do I get for a 9 hour long business trip?}', 
the only relevant node returned by the \ac{LLM} is `\emph{For business trips lasting less than 8 hours, you are not entitled to a per diem}',
which does not answer the question because the user's time requirement is outside the answer's time interval.
This is particularly problematic, as there is another node that answers the question:  `\emph{For trips lasting more than 8 hours, but less than 14, you are entitled to a per diem of 6€.}'.

To improve the model performance and reasoning, we tested the following interventions 1) adding in-context examples to the model prompt and 2) adding an additional reasoning step by requiring the model to also output its justification for marking facts as relevant.

In the ablation study in table \ref{tab:resources}, we can see that both the in-context examples and additional reasoning step result in improved recall.
While justifications slow the model down by generating more output tokens, we consider the improvement in recall as more important, as retrieval errors  define the upper bound on task performance.

Based on these experiments, we see that combining pre-filtering, justifications and in-context examples provides the best recall, and a good trade-off in response time and resource usage.

\subsection{RQ3: Performance In Simulation}

\begin{table*}[htb]  
        \begin{tabular}{|l|l|c|c|c|c|}
        \hline
        \textbf{Domain} & \textbf{Dialog Policy} & \textbf{\makecell{Dialog Success\\(\%)}} & \textbf{\makecell{Dialog Length\\(Guided)}} & \textbf{\makecell{Dialog Length\\(Free)}} & \textbf{\makecell{Interaction\\Mode (F1)}} \\
        \hline
                    & CTS-RL  		& 73.86 & 13.56 & 2.95 & 0.94 \\
        REIMBURSE-en   & CTS-LLM (Gemma-2)    & 77.00 & 10.79 & 2.91 & 0.89 \\
                    & CTS-LLM (GPT-4o-mini) & \textbf{84.20**} & 10.31 & \textbf{2.79} & \textbf{0.95} \\\hline
        
                    & CTS-RL 		   & 76.31 & 6.42 & \textbf{2.29} & n/a \\
        DIAGNOSE    & CTS-LLM (Gemma-2)     & 94.60** & 5.24 & 3.36 & \textbf{1.0} \\
                    & CTS-LLM (GPT-4o-mini) & \textbf{98.80**} & 5.26 & 3.52 & \textbf{1.0} \\
        \hline
        
                    & CTS-RL 		& 73.61 & 7.88 & 2.98 & n/a \\
        ONBOARD     & CTS-LLM (Gemma-2)     & 95.00** & 4.94 & \textbf{2.96} & \textbf{1.0} \\
                    & CTS-LLM (GPT-4o-mini)  & \textbf{96.00**} & 4.88 & \textbf{2.96} & 0.99 \\
        \hline
        
        \end{tabular}
    \centering
    \caption{Objective performance metrics of our proposed CTS-LLM agent compared to the \ac{RL}-based \ac{CTS} baseline \citep{vathZeroDataControllableAdaptive2024} ** represents significant increases ($p<0.0001$ Barnard Exact test). We bold the best results per model and domain, except for guided dialog length, where the goal is to obtain context, rather than simply efficient navigation.}
    \label{tab:dialogs:simulation}
\end{table*}

Evaluating dialog performance against a simulated user is common practice in the dialog community \citep{ai-weng-2008-user}, where it is expensive to test development systems on real users.
To this end, the \ac{CTS} task provides a user simulator to evaluate the system's performance \cite{vathConversationalTreeSearch2023}.
To simulate a dialog, a random goal is chosen from the dialog graph along with a tree path to reach that goal.
The interaction mode is set to either \emph{free} mode or \emph{guided} mode.
When free mode is chosen, a random user question associated with the goal node is drawn as the first user utterance.
If guided mode is chosen, the simulator will select a paraphrase of one of the pre-defined intents at the start node as first utterance.
For each node along the goal path that accepts user input, a random paraphrase of that intent will be chosen .
If the simulated user reaches their goal, the dialog is rated successful (see Appendix \ref{appendix:user_sim} for more details).
Table \ref{tab:dialogs:simulation} shows performance metrics against the \ac{CTS} simulator when evaluating on 500 simulated dialogs for our \ac{LLM} agent and a  state-of-the-art \ac{RL} agent \citep{vathZeroDataControllableAdaptive2024} on the test splits of all three domains.

Using GPT-4o-mini, our method significantly improves dialog success in all three domains ($p<0.0001$; Barnard Exact test).
From this, we see that in addition to the sustainability advantages  our proposed agent brings by eliminating the need for training, it is also able to outperform the previous \ac{RL}-based state-of-the-art \ac{CTS} agent.
We also see that this performance generalizes across domains.

Additionally, when looking at the results for the Gemma-based agent, we find that even a smaller \ac{LLM}, which can be run locally (e.g. on a single NVIDIA A600), can improve dialog success on all domains compared to the \ac{RL}-based \ac{CTS} agent, with significant improvements in the ONBOARD and DIAGNOSE domains ($p<0.001$; Barnard Exact test).

\subsection{RQ4: Performance With Real Users}
For the user evaluation, we selected the GPT-4o-mini-based policy, as it achieved the highest task success in simulation. 
To compare it with a state-of-the-art \ac{RL}-based \ac{CTS} agent, we use the results reported for the REIMBURSE-en domain in \citet{vathZeroDataControllableAdaptive2024} and conduct a parallel user study with our LLM-based agent.
As no user studies exist for the \ac{RL} agent on other domains, recruiting sufficient participants to evaluate both agents across multiple domains would have been prohibitively expensive.
Thus, we focus on the challenging real-world REIMBURSE-en domain in this study.
Example interactions can be found in Appendix \ref{sec:app:dialog_examples}.

\begin{table}[htb] 
        \begin{tabular}{|l|c|c|c|c|}
        \hline
        \textbf{Policy} & \makecell{\textbf{Success} \\(\%)}& \makecell{\textbf{Avg.} \\ \textbf{Turns}} & \textbf{\makecell{Subj.\\Length}} & \textbf{\makecell{Subj.\\Quality}} \\
        \hline
        CTS-RL  & 77.05 & \textbf{7.38} & \textbf{2.92} & 2.87 \\
        \hline
        CTS-LLM & \textbf{86.76*} & 10.24 & 2.75 & \textbf{2.94} \\
        \hline
        \end{tabular}
    \centering
    \caption{User study results for objective and subjective dialog metrics for CTS-RL \citep{vathZeroDataControllableAdaptive2024} and our method. * represents significant differences ($p<0.05$). Perceived quality is a 4-point scale (1 = not answered; 4 = fully answered), perceived length a 5-point scale (1 = much too short; 5 = much too long).}
\label{tab:user_study}
\end{table}

Table \ref{tab:user_study} shows the results of our policy compared to the results of the \ac{RL}-based \ac{CTS} agent.
Performing a Barnard-Exact test, we find that our method led to a significant increase in dialog success compared to the \ac{RL}-based \ac{CTS}  agent ($p < 0.05$) without causing any negative effects in how users perceived the quality of answers they received.
We additionally find that although the CTS-LLM dialogs were objectively longer, the subjective length of the system was rated similarly (even slightly shorter) to the \ac{RL}-based \ac{CTS} agent.
The shorter paths from the \ac{RL} agent are likely explained by the optimization goal, which optimizes dialog length.
Thus, a future goal is to improve the dialog length in the planning step.

\begin{table}[htb] 
        \begin{tabular}{|l|c|c|c|}
        \hline
        \textbf{Policy} & \textbf{Trust} & \textbf{Reliability} & \textbf{Usability} \\
        \hline
        CTS-RL  & \textbf{3.16} & 2.96 & 62.83 \\
        \hline
        CTS-LLM & 3.00 & \textbf{3.06} & \textbf{67.90} \\
        \hline
        \end{tabular}
    \centering
    \caption{User study results for perceived trust, reliability and usability. No significant differences were found between CTS-RL \citep{vathZeroDataControllableAdaptive2024} and our results.}
\label{tab:user_study_trust}
\end{table}

When looking at the subjective perception of the agent in table \ref{tab:user_study_trust}, we find that users rated both reliability and usability of our proposed method higher.

\section{Conclusion}

In this paper, we explore how to apply the reasoning and zero-shot capabilities of \acp{LLM} to dialog planning in a way that allows steering the conversation towards an overarching goal, while at the same time avoiding hallucination and retaining an expert-controllable dialog flow.
We introduce a novel task-oriented dialog policy that performs dialog planning based on a combination of a fast Sentence Transformer, \acp{LLM}, and graph algorithms.
Our dialog policy conforms to the controllability aspect of the \ac{CTS} task by using \acp{LLM} to help with path planning, rather than generating output for the user, avoiding the possibility of hallucination.
This approach also removes the need for training \ac{CTS} dialog policies, allowing for instantaneous adaptation to changes in domain graphs.

Although \ac{RL} is known for its planning capabilities, especially in the task-oriented dialog policy setting, we demonstrate that our \ac{LLM}-based approach to dialog planning can significantly outperform state-of-the art \ac{RL}-based dialog policies.
These results hold across all three domains tested in simulation, and also translate to evaluation on the REIMBURSE-en domain with real users.

In addition, we identify computational resource and reasoning challenges, and demonstrate working solutions to speed up response time and reduce the number of input and output tokens.
In this way, we improve policy performance and effectively decouple the slow filtering step from the graph size.

Finally, our simulations demonstrate that our approach yields good results not only using very large commercial \ac{LLM}, but also when using a smaller open-source \ac{LLM} that can be run on a single GPU.
This allows for more data safety and less costly experimentation and deployment.

\appendix

\section{Limitations}
Although we list the exact version numbers of the \acp{LLM} we use, commercial \acp{LLM} might get updated, thus impacting the exact reproducibility of our results obtained using commercial \acp{LLM}.

Additionally, at the time of writing, there are only three domains available for the \ac{CTS} task.
Therefore, we can evaluate our proposed method only on these domains.
The evaluation with real users is limited to a single domain due to cost reasons.

As discussed in the results, the \ac{LLM} policy interactions, although perceived shorter than the \ac{RL} policy interactions in the user study, are objectively several turns longer than the \ac{RL} policy interactions.

Although we save the computational resources required to train \ac{CTS}-\ac{RL} agents, the runtime resources of the \ac{CTS}-\ac{LLM} are higher.

\section{LLM Prompts}
\label{sec:llm_prompts}

\subsection{Interaction Mode Classification}

Table \ref{tab:appendix:prompts:interaction_mode_classifier} shows the full prompt for the interaction mode classifier.

\begin{table}[h!]
    \centering
    \begin{tabularx}{\linewidth}{c|X}
        \textbf{Role} & \textbf{Prompt} \\ \hline
        variables & \emph{user\_utterance} \\ \hline
        system & Answer with "yes" or "no" only. \\
        user & Is the following text a question / command / requesting or not: \emph{user\_utterance}?
    \end{tabularx}
    \caption{Prompt for classifying which interaction mode (guided or free) the user wants.}
    \label{tab:appendix:prompts:interaction_mode_classifier}
\end{table}

\subsection{Intent Classification}

Table \ref{tab:appendix:propmpts:intent_classifier} shows the full prompt for the user intent classifier.

\begin{table}[h!]
    \centering
    \begin{tabularx}{\linewidth}{c|X}
        \textbf{Role} & \textbf{Prompt} \\ \hline
        variables & \emph{user\_utterance}, \emph{intent\_candidates} \\ \hline
        system & Given this list of possible response candidates: 
        \newline \emph{intent\_candidates} \newline
        Decide which of the response candidate texts most closely matches the user intent, and only output the responses index. Do not output any other text, any code, or anything else. \\
        user & \emph{user\_utterance}
    \end{tabularx}
    \caption{Prompt for classifying which pre-given intent the user is most likely trying to choose.}
    \label{tab:appendix:propmpts:intent_classifier}
\end{table}

\subsection{Goal Node Filter}

\begin{table}[h]
    \centering
    \begin{tabularx}{\linewidth}{c|X}
        \textbf{Role} & \textbf{Prompt} \\ \hline
        variables & \emph{user\_utterance}, \emph{node\_texts}, \emph{examples} \\ \hline
        system & 
    You will be provided with a json list of facts and a query.
    You are to act as a first filter to decide which of the given facts answer the query or are relevant to answering the query, at least partially, and which ones are not relevant to answering the query at all?
    Assign each fact a relevance indicator between 0 and 2, and add a justification of why it is relevant (2), partially related (1), or irrelevant (0).
    Facts are also considered relevant if they imply the answer.
    If facts contain placeholders inside curly braces, assume the placeholder will be filled with a reasonable value.
    Don't return anything besides the json list of relevant facts, and only return facts with relevance indicator higher than 0. Don't return code or additional text.

    REMEMBER: even if some facts are only slightly relevant to answering the query, it is better to rate them with a relevance of 1 than to have all facts have relevance 0.

    \emph{examples}

    Note that the fact with key 4 was excluded from the output, as it has a relevance of 0: Fact 4 is not related to the query about the weather in Singapore. \\
        user & ======= Facts =======\newline
        \emph{node\_texts} \newline
        ======= Query ======= \newline
        \emph{user\_utterance}
    \end{tabularx}
    \caption{Prompt for filtering to keep only relevant goal node candidates}
    \label{tab:appendix:prompts:goalfilter}
\end{table}

Table \ref{tab:appendix:prompts:goalfilter} shows the full prompt for the goal node filtering step.
The variable \emph{examples} is filled with the following text:

\begin{quote}
\lstdefinelanguage{json}{
  basicstyle=\ttfamily,
  showstringspaces=false,
  breaklines=true,
  string=[s]{"}{"},
  comment=[l]{:\ "}, 
}

For example, given the facts:
\begin{lstlisting}[language=json,breaklines=true]
[{"key": 0, "fact": "In Singapore, at 9 a.m., it is usually around 35 degrees celsius."},
{"key": 1, "fact": "In Singapore, between 8 a.m. and 11 a.m., the weather is around 35 degrees celsius."},
{"key": 2, "fact": "In London, at 9 a.m., it is usually 25 degrees celsius."},
{"key": 3, "fact": "In Singapore, between 10 a.m. and 11 a.m., it is usually around 30 degrees celsius."},
{"key": 4, "fact": "In Singapore, there are many tourist attractions."},
{"key": 5, "fact": In Singapore, it is usually around 35 degrees celsius in the mornings, but cooler in the evenings."},
{"key": 6, "fact": "In {{ COUNTRY }}, at 9 a.m., it is usually around 35 degrees celsius."}]
\end{lstlisting}

And a query:
"What is the weather usually in Singapore at 9 a.m.?"
The reply should only be a json list of the facts, indicating if the facts are related to or directly answering the query, formatted like this:

\begin{lstlisting}[language=json,breaklines=true]
[{"key": 0, "relevance": 2, "justification": "The fact is relevant because it answers the user request perfectly"},
{"key": 1, "relevance": 2, "justification": "The fact is relevant as it answers the user request, because the requested time of 9 a.m. lies between the fact's timespan of 8 a.m. t0 11 a.m."},
{"key": 2, "relevance": 1, "justification": "While the time is correct, the fact is listing the temperature for London instead of Singapore"},
{"key": 3, "relevance": 1, "justification": "The fact is talking about the weather in Singapore, which is relevant to the user, although the requested time of 9 a.m. lies outside the fact's timespan of 10 a.m. to 11 a.m."},
{"key": 5, "relevance": 2, "justification": "The fact is relevant as it partially answers the user query: while it does not state a specific time, it implies the temperatures in Singapore at the requested time"},
{"key": 6, "relevance": 2, "justification": "The fact is relevant as it could answer the user request perfectly, once the placeholder is filled."}]
\end{lstlisting}
\end{quote}

\section{User Simulator}
\label{appendix:user_sim}

All dialog success scores and conversation lengths reported from simulated users are the average across 500 dialogs.

At the start of each simulated dialog, a goal node is drawn randomly from the domain graph.
Each interaction then begins at the (unique) start node, which is output to the user.
Here, the initial user utterance is generated depending on the dialog mode.
In free mode, a question associated with the goal node is drawn randomly.
In guided mode, a paraphrase for the user intent on the path to the goal is randomly chosen instead.
Control is then handed over to the policy.
When the policy chooses to output a node to the user that requires a user response, the user response is randomly chosen from a list of available paraphrases associated with a user intent that facilitates reaching the goal.
A dialog ends once the goal was reached, the user patience was reached (encountering the same dialog node multiple times), or the maximum turn length was reached (four times the maximum tree depth).
To support the training of \ac{RL} agents, each turn is stored in an experience buffer.
The simplified pseudo-code for the user simulator is shown in listing \ref{alg:cts:simulator_new}.
For more details, we refer to \citet{vathConversationalTreeSearch2023} and \citet{vathZeroDataControllableAdaptive2024}.

\begin{algorithm}[h!]
\caption{Pseudo code for the user simulator, adapted from \citet{vathConversationalTreeSearch2023} and \citet{vathZeroDataControllableAdaptive2024}.}
\label{alg:cts:simulator_new}
\KwData{$\text{questions} \gets \text{List of Questions associated with dialog tree nodes}$}
\KwData{$\text{answers} \gets \text{List of Answer synonyms}$}
\KwData{$N \gets \text{Number of dialogs to be simulated}$}
\KwData{$G(V,E)$ \tcp*[r]{Dialog Tree}}
\KwData{$T \gets \text{Maximum number of turns per dialog}$}

$n \gets 0$\;
\For{$n < N$}{
    $v \gets \text{start node}$\;
    $m \gets random\_uniform(\{ \text{Free}, \text{Guided} \})$\;
    $g \gets random\_uniform(\{ g \in V:  |questions(g)| \geq 1 \})$\; 
    $p \gets (e(v,v_1), \dots, e(v_N,g)) \subset E$\;
    $u \gets random\_uniform(questions(g))$\;
    $t \gets 1$\;
    \While{$v \neq g \wedge t \leq T \wedge \text{neighbors}(v) \neq \emptyset$}{
        $a \gets policy(v, u)$ \tcp*[r]{Select action}
        \eIf{$a = \text{ASK(v)}$}{
            $print(v)$\;
            $e \gets e(v,v')$ \tcp*[r]{$v'$ is $g$ (guided mode) or next node after $v$ (free mode)}
            \eIf{$type(v) = \text{information}$}{
                $u \gets \emptyset$\;
            }{
                $u \gets random\_uniform(questions(e))$
            }
        }{
            \If{$a = \text{SKIP(v,v')}$}{
                $v \gets v'$\;
                $u \gets \emptyset$\;
            }
        }
        $r \gets \text{Calculate rewards}$\;
        Store $a, v', r$ in replay buffer\;
        $t \gets t + 1$\;
    }
    $n \gets n + 1$\;
}
\end{algorithm}

\section{Example Interactions}
\label{sec:app:dialog_examples}
In this section we provide examples of different types of interactions users had with our \ac{LLM}-based \ac{CTS} agent.

In figure \ref{fig:dialog_ex1}, we see an example of a user who had a concrete question and was given a concrete answer.
In this case, our agent found two possible goals along the same branch in the dialog graph that might answer the user's query, as there were no branches between the two answers which could determine which was more relevant, both were output, leading to a more complete answer being given to the user.

\begin{dialogue}
\speak{GOAL NODE} (16460328708250870): If you are also transporting other coworkers, you are entitled to 2 Cent per kilometer.

\speak{USER INSTRUCTIONS} 
You want to figure out how much money you can get for your travel.  \\
1. You used your own car.  \\
2. You took two colleagues with you. 

\speak{SYSTEM} (NODE: 16348058621438633)  
What topic do you have questions about?  
You can either click on an answer from the suggested topics or enter your own text.  

\speak{USER} Hi, on my recent trip I took my own car and carried two colleagues, how much money can I claim back for this please?  

\speak{SYSTEM} (NODE: 16460325754790153)  
If you used your car for a pressing reason, you are entitled to 0.25€ per kilometer driven.  

\speak{SYSTEM} (NODE: 16460328708250870)  
If you are also transporting other coworkers, you are entitled to 2 Cent per person and kilometer for each passenger who is also eligible for travel reimbursement.  
(To claim a passenger, they must also sign your travel reimbursement request.)  
If you are transporting business-related items (e.g., files) weighing more than 50 Kg, you are eligible for 2 Cent per 50 Kg and kilometer.  
(Changes coming in 2022)  

\speak{DIALOG END} SUCCESS  \\
SUBJECTIVE QUALITY: 3  \\
SUBJECTIVE LENGTH: 2  
\end{dialogue}
\begin{center}
  \captionsetup{type=figure}
  \caption{Example dialog of a direct question which is given a direct answer. Excerpt from the conversations with user \emph{c7bb706cdb623739bf9e80c638e4eb}.}
  \label{fig:dialog_ex1}
\end{center}

In figure \ref{fig:dialog_ex2}, we again see a user asking a direct question, however in this case, the system has two possible answer candidates (train or plane seat reservations) and needs to ask the user to clarify which of these is correct in order to answer.
Once the user does this, there is only one answer candidate left and this is directly output to the user.

\begin{dialogue}
\speak{GOAL NODE} (16363756478730906): Seat reservations are allowed for train travel. 
\speak{USER INSTRUCTIONS}
You want to know if you can get reimbursed if you reserve a seat for yourself on the train.  

\speak{SYSTEM} (NODE: 16348058621438633)  
What topic do you have questions about?  
You can either click on an answer from the suggested topics or enter your own text.  

\speak{USER} I want to book a seat on the train. Can I get a refund if needed?  

\speak{SYSTEM} (NODE: 16348100109577394)  
What type of transportation would you like?  

\speak{USER} Train  

\speak{SYSTEM} (NODE: 16363756478730906)  
Seat reservations are allowed for train travel.  

\speak{DIALOG END} SUCCESS \\
SUBJECTIVE QUALITY: 2  \\
SUBJECTIVE LENGTH: 3  

\end{dialogue}
\begin{center}
  \captionsetup{type=figure}
\caption{Example dialog of a direct question which needs clarification before being answered.
Excerpt from the conversations with user \emph{6248a541d241dda06d678a24a511ce}.}
\label{fig:dialog_ex2}
\end{center}

In figure \ref{fig:dialog_ex3}, we see a different type of dialog interaction.
Here the user does not formulate a direct question and instead chooses one of the offered general topics.
From here, the agent guides the user through each node in that branch of the dialog tree, giving them additional context and asking clarifying questions until they get the answer to their question.

\begin{dialogue}

\speak{GOAL NODE} (16370483534787100): Your business trip must be planned as though you were traveling alone.  

\speak{USER INSTRUCTIONS}: You want to know how the reimbursement process works for a research semester.   \\
1. You plan to bring your family with you. 

\speak{SYSTEM} (NODE: 16348058621438633)  
What topic do you have questions about?  
You can either click on an answer from the suggested topics or enter your own text.  

\speak{USER} Research semester  

\speak{SYSTEM} (NODE: 16370487978253782)  
How long will your research semester last?  

\speak{USER} 2 days  

\speak{SYSTEM} (NODE: 16353297472197738)  
Is this as part of a reallocation?  

\speak{USER} No  

\speak{SYSTEM} (NODE: 16370501863636503)  
As there has been no reallocation, your research semester counts as a business trip.  

\speak{SYSTEM} (NODE: 16353303012891903)  
Are you traveling alone or with your family?  

\speak{USER} With family  

\speak{SYSTEM} (NODE: 16370483534787100)  
Your business trip must be planned as though you were traveling alone (without your family).  

How does that work?  
1. Provide receipts for only the costs associated with purchases for yourself.  
2. Collect comparison offers where necessary (e.g., transportation costs and accommodation) for what the costs would have been for only one person.  
3. Personal expenses must be strictly separated from business expenses.  

You are responsible for paying all additional costs that result from bringing your family along.  

\speak{DIALOG END} SUCCESS  \\
SUBJECTIVE QUALITY: 1  \\
SUBJECTIVE LENGTH: 2  

\end{dialogue}
\begin{center}
  \captionsetup{type=figure}
\caption{Example dialog of a guided mode interaction, where the user does not ask a direct question and must be asked follow-up questions to clarify their intent.
Excerpt from the conversations with user \emph{c017850ef3e9fd55350d0c8374859f}.}
\label{fig:dialog_ex3}
\end{center}

\section*{Ethical Considerations and Risks}
To ensure that users could give informed consent, we provided a detailed description of the task and research objectives both on the crowdsourcing platform and once they had accepted the task.
In respect of participant privacy, we specifically did not collect personally identifying data from any users.
To this end, we store all logs and survey responses using an anonymous hash generated based on a given username, rather than with the username itself.
In this way, users could log in again if they needed to take a break in the middle of the interaction, but we had no way of directly linking any recorded results to, e.g., users' Prolific account identifiers.
To ensure that participants were fairly compensated, we followed best practices recommended by the crowdsourcing platform paying users at 12.60€/hr, which was in-line with minimum wage in the country of our research institution.
We additionally used our pilot study to verify that our estimated time was below the median time we selected when advertising the task, meaning most participants had a higher hourly wage.

In terms of risks, the goal of this paper is to create more effective dialog agents with zero training effort.
While we try to mitigate the possibility of chatbots generating false information, our work has the possible risk of creating chatbots that could also be used to replace human jobs.

\bibliography{aaai25}

\end{document}